# Triage Accuracy of Jev and Open-Weight System One Decision Models and Their Use for Early Emergency Recognition in Symptom-Assessment Applications

Marvin Kopka[1,2,3*], Austen El-Osta[3], Christoph Kollwitz[1], Felix Balzer[2] and Florian Bontrup[1,4]

[1] DOCYET GmbH, Leipzig, Germany

[2] Institute for Medical Informatics, Charité-Universitätsmedizin Berlin, corporate member of Freie Universität Berlin and Humboldt-Universität zu Berlin, Berlin, Germany

[3] Self-Care Academic Research Unit (SCARU), Department of Primary Care and Public Health, Imperial College London, London, United Kingdom

[4] Institute of Information Systems, Leuphana University, Lüneburg, Germany

[*] Correspondence

## Abstract

Laypeople increasingly use large language models (LLMs) for triage advice, but LLMs answer in variable free text, which makes it hard to evaluate them for use in medical devices. Decision models, which assign a probability to each predefined triage category, may be easier to evaluate. To test their triage accuracy, we evaluated 16 decision models (14 open-weight, 2 proprietary) on 233 synthetic and 10,000 real-world cases, yielding 163,728 total assessments. The best models triaged 72.1% of synthetic cases correctly and agreed with an existing algorithm in 56.1% of real-world cases. They were more promising for recognising emergencies early: given only demographics, chief complaints, risk factors and red-flag answers, the best model reached an AUROC of 0.87. At a 5% error margin, it could decide for or against emergencies for about half of users. Decision models should therefore not replace triage engines but could shorten symptom assessments at setting-specific error margins.

# Introduction

Deciding whether and how urgently to seek care is one of the most common health decisions [1,2], and laypeople increasingly make their decision with the help of large language models (LLMs) [3,4]. In a US survey, about half of those who used ChatGPT for health information did so to decide whether to see a health professional [3], whereas in Australia more than a third of ChatGPT health users assessed their symptoms with it [4]. Since this advice can directly change whether and where people seek care, errors can be harmful in two ways: advising users to self-care in an emergency directly endangers them (“undertriage”), whereas advising almost everyone to seek care (“overtriage”) could result in overuse of limited resources disenfranchising other patients [5].

Because of these risks, the EU Medical Device Regulation (MDR) classifies software that supports triage and diagnostic decisions as a medical device of at least class IIa, which requires manufacturers to demonstrate its safety and performance before market entry [6]. Dedicated symptom-assessment applications (SAAs), which typically use transparent algorithms such as Bayesian networks or rule-based systems [7–9], are certified accordingly. LLMs that give medical advice to laypeople technically may fall under the same rules, but are often available without medical device certification [10,11]. Judged by accuracy alone, LLMs would not perform worse than certified SAAs: they reach about 58–76% accuracy, which falls within the wide range of SAAs (12–90%) [5,12]. Unlike SAAs, however, which tend to overtriage [5,7,12], LLMs seem to make both types of error described above: they rarely recommended self-care and undertriaged more than half of true emergencies in two recent evaluations [13,14].

Even if these errors were reduced, approving LLMs would remain difficult for two reasons. First, they generate free text instead of one of the mandated triage levels, such as those specified in Germany’s draft act on data and digital innovation in healthcare (GeDIG) [15], so their recommendations have to be mapped to these levels before they can be used or evaluated. Second, they are non-deterministic, that is, the same input can lead to different outputs. When ChatGPT triaged the same cases repeatedly, identical prompts led to substantial variation, sometimes advising every possible care level for the same case [13,16]. This conflicts with the MDR, which requires software to be designed to ensure repeatability and reliability [6], and complicates evaluation because one run does not show how a model would respond to the same case in the next. A system that is as flexible as an LLM (e.g., accepting free-text input), but returns one of the mandated triage levels and responds consistently to the same case, would thus be easier to integrate into certified software and to evaluate.

Decision models (also called “system one models”) like Jev, a novel class of models released in September 2026, may offer this combination. Instead of generating text, they select one of several response options defined in advance, return a probability for each option, and often respond in less than a second [17]. One of their advantages is that their recommendations change less often than those of LLMs [18,19]. Since several decision models are open-weight with some having only a few billion parameters, manufacturers could also run them locally without transferring health data to external providers [20,21]. Their fixed options and probabilities also allow new designs for symptom assessment: symptoms could be extracted from free text, symptom assessments could be shortened, safety could be improved by identifying emergencies early on, and probability thresholds may be used to adapt recommendations to acceptable error margins.

Whether decision models give accurate triage advice, however, remains unknown. In early evaluations in the medical domain, their performance depended on the task: although they matched frontier LLMs on research questions, they fell behind on complex diagnostic cases [17] and numerical tasks [18]. Results from other (medical) tasks can therefore not be transferred to appraise their triage accuracy.

To address this gap in knowledge, we conducted an exploratory diagnostic accuracy study. Our primary research question was how accurate the triage advice of decision models was. To assess this question, we evaluated the triage accuracy of 14 open-weight and two proprietary decision models, using both synthetic cases from various previous evaluations [9,22–33] and real-world data from an SAA used by laypeople in Germany. We focused on open-weight models because they can be run locally and included two proprietary models as comparators.

Beyond the primary research question, decision models might help with two common problems of symptom assessment in specific sub-decisions, even if they are not accurate enough for the triage decision as a whole: (i) inaccurate and often overcautious advice, and (ii) long assessments. Each problem led to an exploratory research question. First, errors may partly come from choosing one of several triage levels at once, which is more complex than answering a binary question, such as whether a case is an emergency or not. Previous research therefore divided triage decisions into a sequence of binary questions and assessed accuracy on each [25,34]. Because decision models select from predefined options, triage can be posed to them in both ways. Thus, our first exploratory research question was: Does a decision tree of three binary questions change accuracy compared with a direct triage decision? As part of this question, we also assessed the sensitivity and specificity of all models on each sub-decision to locate where over- and undertriage occur. Second, SAAs often ask many questions before giving a recommendation, which makes assessments long [35] and implies that some users in an emergency may receive advice only at the end. If decision models could identify emergencies early, the assessment could be shortened for these users. This led to our second exploratory research question: Can decision models identify emergencies based only on demographic information, the chief complaint, risk factors and red-flag questions, and for how many users can this decision be made early at an accepted error margin? Within this exploratory design, we thus compared not only different models but also different ways of posing the triage task: a direct question with all triage levels, a decision tree of binary questions, and a reduced input for early emergency recognition.

# Methods

## Study Design

This study was designed as a retrospective diagnostic accuracy study of decision models. We evaluated 16 models with 233 synthetic cases and 10,000 real-world cases. The primary outcome was triage accuracy (ranging from 3 to 4 levels), and the secondary outcomes included (a) the degree of over- and undertriage, (b) sensitivity, and (c) specificity. Since the triage levels differed between the data sources, we included a sensitivity analysis that mapped cases with four triage levels to a three-tiered triage level system in line with the SCARF checklist [36]. For the real-world cases, we additionally split the triage decision into binary questions and tested whether emergencies can be identified from a reduced input. All analyses were exploratory. Results are reported according to the STARD and the SCARF checklists [36,37].

## Cases & Index Tests

### Synthetic cases

We used five previously published synthetic case sets that were identified through previous systematic reviews [5,7,12,38]. The first set from Semigran et al. contains 45 cases with three triage levels [22]. Two further sets adapted the original vignettes, added cases and used four triage levels (48 cases each) [23,28]. We retained all cases across these three sets because each was evaluated against the reference standard and triage scale of its source dataset. Another set by Delshad et al., which was unrelated to that of Semigran et al., contained 50 cases with four triage levels, for which the reference standard was determined by a majority vote of physicians [29]. We excluded three of these cases because the vote did not result in a clear triage level, leaving 47 cases in the set. Another unrelated set by Kopka et al. containing 45 cases with four triage levels that are based on real patient questions and written in lay language was also used [24]. All other sets are phrased for clinicians. We did not exclude atypical cases.

### Real-world cases

We extracted a random sample of 10,000 cases that were completed in the DOCYET SAA between 1 January and 20 October 2026. This SAA is used by laypeople in Germany for care-seeking decisions and is integrated in several healthcare settings, including health insurers, general practitioner (GP) practices, and emergency departments. No atypical cases were deliberately excluded. We considered a sample size of 10,000 cases appropriate because it provides sufficient precision while keeping the number of API calls feasible.

### Reference standard

For the synthetic cases, we used the triage levels assigned in the original publications. In all case sets, these were assigned by at least two licensed physicians. For the real-world cases, we used the triage level assigned by the DOCYET SAA prospectively during use as reference standard, because having physicians assess 10,000 cases was deemed not feasible. For these cases, accuracy, sensitivity and specificity therefore express agreement with the SAA rather than clinical correctness. Four triage levels were distinguished: (1) emergency, (2) urgent care, defined as consulting a doctor within 24 hours, (3) non-emergency consultation, and (4) self-care.

For both synthetic and real-world cases, we used the following definitions from previous studies [24,28]:

- Self-care: Let the health issue get better on its own. It most likely doesn't require seeing a doctor.
- Non-emergency: Try to see a doctor within a week. It likely won't get better on its own, but it's also not an emergency.
- Urgent: Try to see a doctor within a day. The issue is urgent, but not an emergency.
- Emergency: Call 112 or go directly to the emergency room. The issue requires immediate attention.

Since the case sets differed in the number of triage levels, we re-queried all models with three triage levels to compare them on the same scale [36]. For this purpose, we merged urgent and non-emergency consultations into non-emergency care and mapped the reference standard accordingly. We took the definitions of the three levels from previous studies [13,28]:

- Self-care: Let the health issue get better on its own. It most likely doesn't require seeing a doctor.
- Non-emergency: Try to see a doctor, but the symptoms do not require immediate attention at an emergency room. For example, you can see a doctor within 24 hours or in the next week.
- Emergency: Call 112 or go directly to the emergency room. The issue requires immediate attention.

## Models

To identify open-weight decision models, we searched the OpenRouter model catalogue on 22 November 2026 and included all open-weight decision models listed there. If several versions of a model were available, we used the most recent one. This search resulted in the inclusion of 14 models: clef, clef-flash, clef-omni, drex-max, bison-1, groundhog-1, elephant-1, joey-1, kev-4b, mammoth-1, pplx-decider-v1.3, pplx-decider-flash-v1.3, span-02 and tev1-4b-experimental. As comparators, we additionally included two proprietary models, gpt-6-luna-decisions and jev-1.14.

## Procedure

The procedure is illustrated in Figure 1. In a first step, all models triaged every case on its original triage scale. As a sensitivity analysis, we repeated this process with three triage levels. In the second step, we only used the real-world cases and tested a decision tree that included several binary questions to arrive at the four triage levels rather than one question with all four triage levels. Lastly, we used the three models that performed best in emergency questions and tested whether they can be used for emergency pre-screening – that is, whether they can make accurate decisions for or against emergency care early in the symptom assessment with limited information.

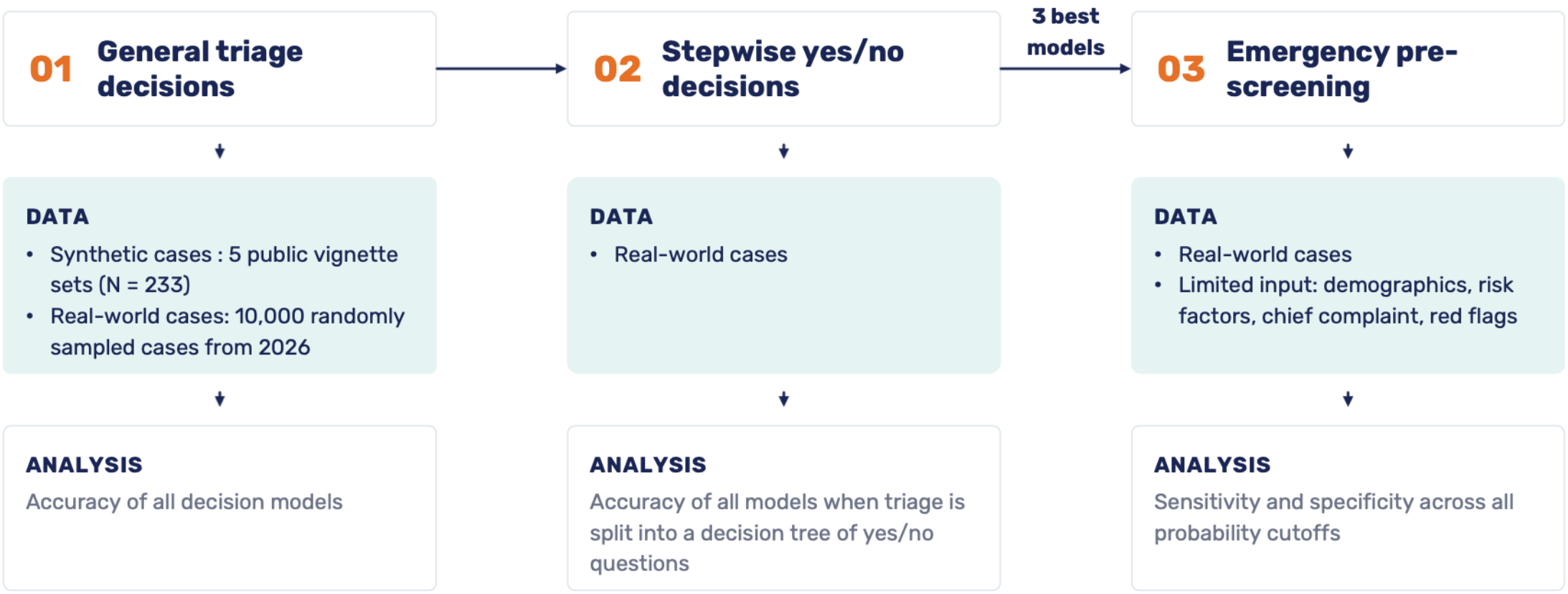

**Figure 1. Data collection and analysis procedure**

We passed each synthetic case in its original text form to the models and used a structured JSON object in German for the real-world cases. Each JSON case contained age and gender, followed by the users' chief complaint, red-flag questions, risk factors and additional follow-up questions on users' symptoms. For each question, it listed whether the user confirmed it, denied it or answered it with "don't know" (see Appendix A for the structure and a fictitious example).

All models were called using the OpenRouter API from a custom-built script in R (version 4.6.1). Since the script entered all cases, no human inputters were involved. For the prompt, we used instructions previously used in evaluating the triage accuracy of regular LLMs [13,24]: "*Please give a triage classification for the following case*", together with the triage levels described in the previous section as response options. The model responses, including the predicted triage level and the class probabilities, were stored in a data frame. Before data collection, we conducted several pretests to ensure that the cases were transferred correctly and that the data were complete. All data were collected on 22 November 2026.

In the second step, we split the triage decision for real-world cases into three binary questions, see Figure 2

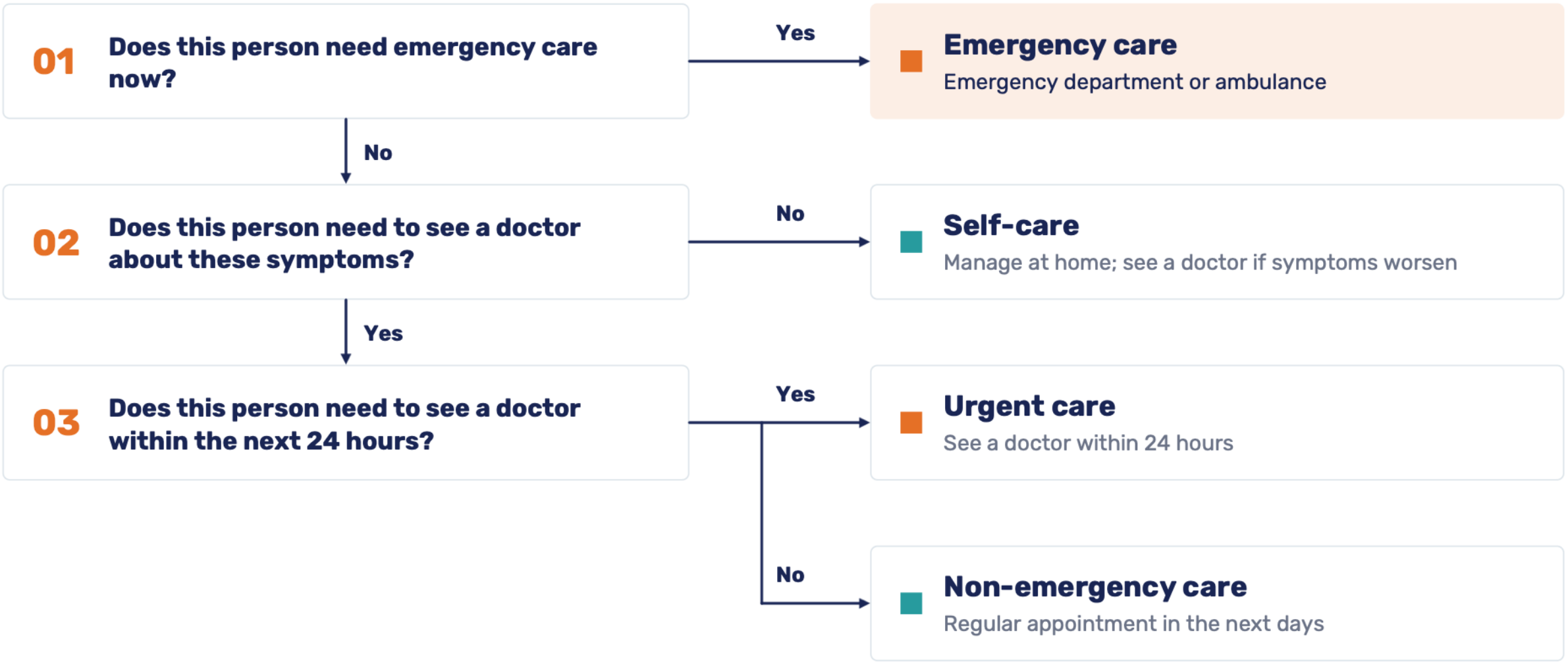

**Figure 2. Decision tree**

The resulting decision tree first asked "*Does this person need emergency care now?*", followed by "*Does this person need to see a doctor about these symptoms, or do these symptoms most likely not require seeing a doctor?*", and "*Does this person need to see a doctor within the next 24 hours?*". Each question had two response options, one of which escalated care. If the first question was answered with the escalating option, the case was triaged as emergency; if the second question was not, it was triaged as self-care. Otherwise, the third question decided between urgent and non-emergency care. As a sensitivity analysis, we again mapped this decision tree to three triage levels (emergency, non-emergency, self-care) by using the first two questions only.

Lastly, we explored whether decision models can make the emergency decision early in the assessment. To this end, we re-queried three open-weight models with a reduced input: the models with the highest sensitivity, the highest specificity, and the highest balanced accuracy on the binary emergency question as identified in the second step. The reduced input contained only the information collected before the follow-up questions: age, gender, risk factors, the chief complaint and the answers to red-flag questions.

## Analysis

Since this was a diagnostic accuracy study, all data were analysed descriptively with confidence intervals (CIs) added as inferential analysis.

First, we calculated accuracy as the proportion of cases in which a model's triage level matched the reference standard, both overall and for each reference level [39]. To assess the direction of errors, we calculated the share of overtriage among all errors and present confusion matrices of model outputs and the reference standard.

Second, for the decision tree, we calculated accuracy as well as sensitivity and specificity for each binary question and compared its accuracy with that of the direct question on the same cases.

Lastly, for emergency recognition, we calculated sensitivity, specificity and balanced accuracy of each model's own decision, and the AUROC based on the predicted probability of an emergency. In a next step, we replaced the model's decision with probability cutoffs, classifying cases below a certain cutoff value as no emergency and above as an emergency. To examine how the choice of cutoff affects triage errors, we plotted the ROC curve of each model and visualised both error types across all cutoff values from 0 to 100%: the share of missed emergencies (1 − sensitivity, i.e., undertriage) and the share of non-emergencies classified as emergencies (1 − specificity, i.e., overtriage). This visualisation allows determining how moving the cutoff reduces one type of error at the expense of the other.

Since a single cutoff cannot keep both errors low simultaneously, we used two cutoffs and divided cases into three groups: cases below a rule-out cutoff (e.g., 1%) were classified as no emergency, cases above a rule-in cutoff as emergency, and all cases in between as uncertain. We set the rule-out cutoff so that at most 1%, 5% or 10% of emergencies were missed, and the rule-in cutoff so that the same share of non-emergencies was classified as emergencies. Cases above the rule-in cutoff would be referred to emergency care immediately and the assessment would terminate. Cases below the rule-out cutoff would have the emergency question settled but would continue with the remaining questions to decide between self-care and non-emergency care. In settings that only require an emergency decision, such as redirection from an emergency department, the assessment could end for these cases as well. This analysis therefore shows for how many users the assessment could be shortened at an accepted error margin. Since different samples can lead to different probability cutoffs, we evaluated both cutoffs with 10-fold cross-validation: cutoffs were derived on nine folds and applied to the held-out fold, and all shares were pooled over the held-out cases. CIs for the cross-validated shares were calculated with the Clopper-Pearson method.

# Results

## Regular Triage Decisions

### Synthetic Cases

We evaluated 16 decision models with 233 synthetic cases from five case sets (four sets with four triage levels and one set with three triage levels), yielding 3,728 assessments.

When each case was triaged on its original triage scale (i.e., three or four options), mammoth-1, an open-weight model, achieved the highest accuracy, correctly triaging 72.1% (168/233; 95% CI 65.9–77.8%) of all cases, followed by the best proprietary model, gpt-6-luna-decisions (71.2% (166/233; 95% CI 65.0–77.0%)). The lowest accuracy was observed for tev1-4b-experimental (52.8% (123/233; 95% CI 46.2–59.3%)). The accuracy of all models on each case set separately can be found in Appendix Table S1.

The highest emergency accuracy was observed for span-02 (91.8% (56/61; 95% CI 81.9–97.3%)), the highest accuracy among urgent cases for clef-flash (82.5% (33/40; 95% CI 67.2–92.7%)), among non-emergency cases for bison-1 (88.5% (69/78; 95% CI 79.2–94.6%)), and among self-care cases for mammoth-1 (61.1% (33/54; 95% CI 46.9–74.1%)). All models tended to overtriage, although the proportion of overtriage among all errors ranged from 52.9% (37/70; 95% CI 40.6–64.9%) for drex-max to 82.7% (67/81; 95% CI 72.7–90.2%) for span-02, see Table 2. The most frequent misclassification was self-care cases triaged as non-emergency (49.3% of all self-care assessments; 426/864), and no model ever triaged emergency cases as self-care, see Figures S1–S5 in the Appendix.

When all cases were triaged with the same three options (self-care, non-emergency, emergency) as a sensitivity analysis, accuracy ranged from 60.5% (141/233; 95% CI 53.9–66.8%) for tev1-4b-experimental to 80.3% (187/233; 95% CI 74.6–85.2%) for mammoth-1. Again, the models tended to overtriage (share of overtriage among all errors 60.9–91.4%), see Appendix Table S2. Self-care cases triaged as non-emergency remained the most frequent misclassification (68.9%; 595/864), see Figures S6–S9 in the Appendix.

**Table 2. Triage accuracy by model on all synthetic cases (n = 233)**

| Model | Overall | Emergency | Urgent | Non-emergency | Self-care | Overtriage share of errors |
|---|---|---|---|---|---|---|
| mammoth-1 | 72.1% (168/233) | 80.3% (49/61) | 52.5% (21/40) | 83.3% (65/78) | 61.1% (33/54) | 58.5% (38/65) |
| gpt-6-luna-decisions | 71.2% (166/233) | 90.2% (55/61) | 55.0% (22/40) | 78.2% (61/78) | 51.9% (28/54) | 65.7% (44/67) |
| bison-1 | 70.8% (165/233) | 72.1% (44/61) | 75.0% (30/40) | 88.5% (69/78) | 40.7% (22/54) | 60.3% (41/68) |
| drex-max | 70.0% (163/233) | 75.4% (46/61) | 60.0% (24/40) | 83.3% (65/78) | 51.9% (28/54) | 52.9% (37/70) |
| pplx-decider-v1.3 | 69.5% (162/233) | 72.1% (44/61) | 75.0% (30/40) | 84.6% (66/78) | 40.7% (22/54) | 60.6% (43/71) |
| groundhog-1 | 67.4% (157/233) | 77.0% (47/61) | 67.5% (27/40) | 80.8% (63/78) | 37.0% (20/54) | 61.8% (47/76) |
| jev-1.14 | 65.7% (153/233) | 83.6% (51/61) | 67.5% (27/40) | 80.8% (63/78) | 22.2% (12/54) | 80.0% (64/80) |
| span-02 | 65.2% (152/233) | 91.8% (56/61) | 62.5% (25/40) | 78.2% (61/78) | 18.5% (10/54) | 82.7% (67/81) |
| clef | 64.8% (151/233) | 75.4% (46/61) | 72.5% (29/40) | 71.8% (56/78) | 37.0% (20/54) | 63.4% (52/82) |
| pplx-decider-flash-v1.3 | 63.9% (149/233) | 80.3% (49/61) | 60.0% (24/40) | 74.4% (58/78) | 33.3% (18/54) | 66.7% (56/84) |
| clef-flash | 59.2% (138/233) | 78.7% (48/61) | 82.5% (33/40) | 67.9% (53/78) | 7.4% (4/54) | 81.1% (77/95) |
| kev-4b | 58.8% (137/233) | 75.4% (46/61) | 75.0% (30/40) | 67.9% (53/78) | 14.8% (8/54) | 75.0% (72/96) |
| elephant-1 | 58.8% (137/233) | 70.5% (43/61) | 67.5% (27/40) | 61.5% (48/78) | 35.2% (19/54) | 62.5% (60/96) |
| clef-omni | 57.5% (134/233) | 67.2% (41/61) | 72.5% (29/40) | 70.5% (55/78) | 16.7% (9/54) | 72.7% (72/99) |
| joey-1 | 54.1% (126/233) | 86.9% (53/61) | 12.5% (5/40) | 62.8% (49/78) | 35.2% (19/54) | 65.4% (70/107) |
| tev1-4b-experimental | 52.8% (123/233) | 82.0% (50/61) | 60.0% (24/40) | 53.8% (42/78) | 13.0% (7/54) | 81.8% (90/110) |

## Real-World Cases

Each model triaged all 10,000 real-world cases, yielding 160,000 total assessments.

When the real-world cases were triaged with all four triage levels, bison-1, an open-weight model, showed the highest agreement with the SAA reference standard, matching its triage level in 56.1% (5,609/10,000; 95% CI 55.1–57.1%), compared to 55.3% (5,526/10,000; 95% CI 54.3–56.2%) for the best proprietary model, gpt-6-luna-decisions. The lowest agreement was observed for joey-1 (22.6% (2,255/10,000; 95% CI 21.7–23.4%)).

The highest agreement with the SAA among emergency cases was observed for joey-1 (93.7% (1,179/1,258; 95% CI 92.2–95.0%)), among urgent (within 24 h) cases for tev1-4b-

experimental (83.0% (1,384/1,667; 95% CI 81.1–84.8%)), among non-emergency cases for clef (83.2% (3,746/4,505; 95% CI 82.0–84.2%)), and among self-care cases for elephant-1 (44.7% (1,150/2,570; 95% CI 42.8–46.7%)). Fifteen of 16 models tended to overtriage rather than undertriage; the share of overtriage among all errors ranged from 46.5% (2,044/4,397; 95% CI 45.0–48.0%) for drex-max to 97.5% (7,553/7,745; 95% CI 97.1–97.9%) for joey-1, see Figure 3. The most frequent misclassification was self-care cases triaged as non-emergency (51.6% of all self-care assessments; 21,236/41,120), and the least frequent misclassification was emergency cases triaged as self-care (0.6% of all emergency assessments; 126/20,128), see Figure 4.

When the models were re-queried with three options (self-care, non-emergency, emergency) as a sensitivity analysis, agreement with the SAA ranged from 28.1% (2,805/10,000; 95% CI 27.2–28.9%) for joey-1 to 70.3% (7,033/10,000; 95% CI 69.4–71.2%) for pplx-decider-v1.3. Again, all models tended to overtriage (with a share of overtriage among all errors from 54.0 to 99.0%), see Appendix Table S3. Self-care cases triaged as non-emergency remained the most frequent misclassification (73.8%; 30,359/41,120), see Figure S10.

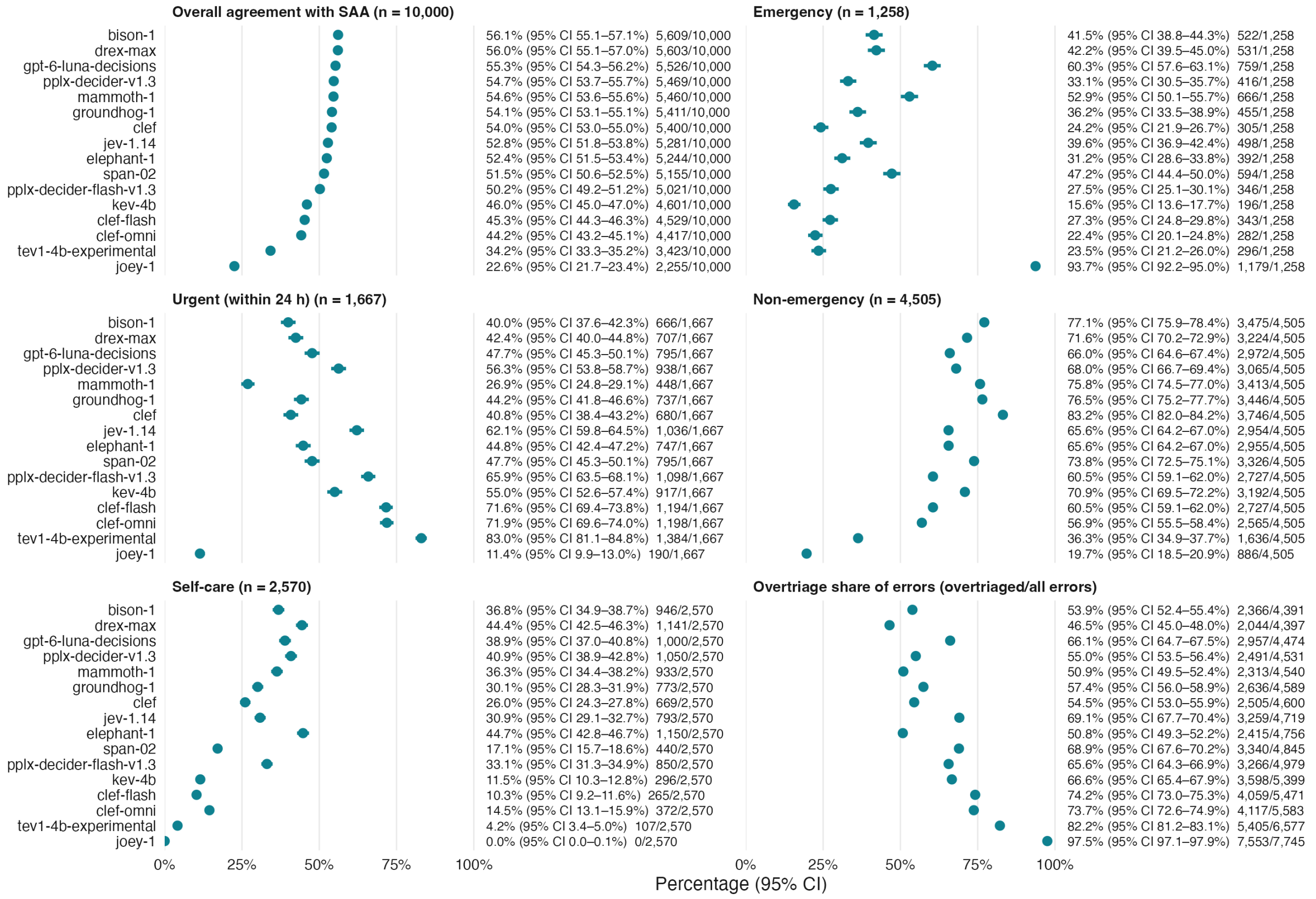

*Note.* Points show agreement with the SAA and overtriage share of errors, lines the 95% confidence interval.

**Figure 3. Agreement with the SAA per model (real-world cases)**

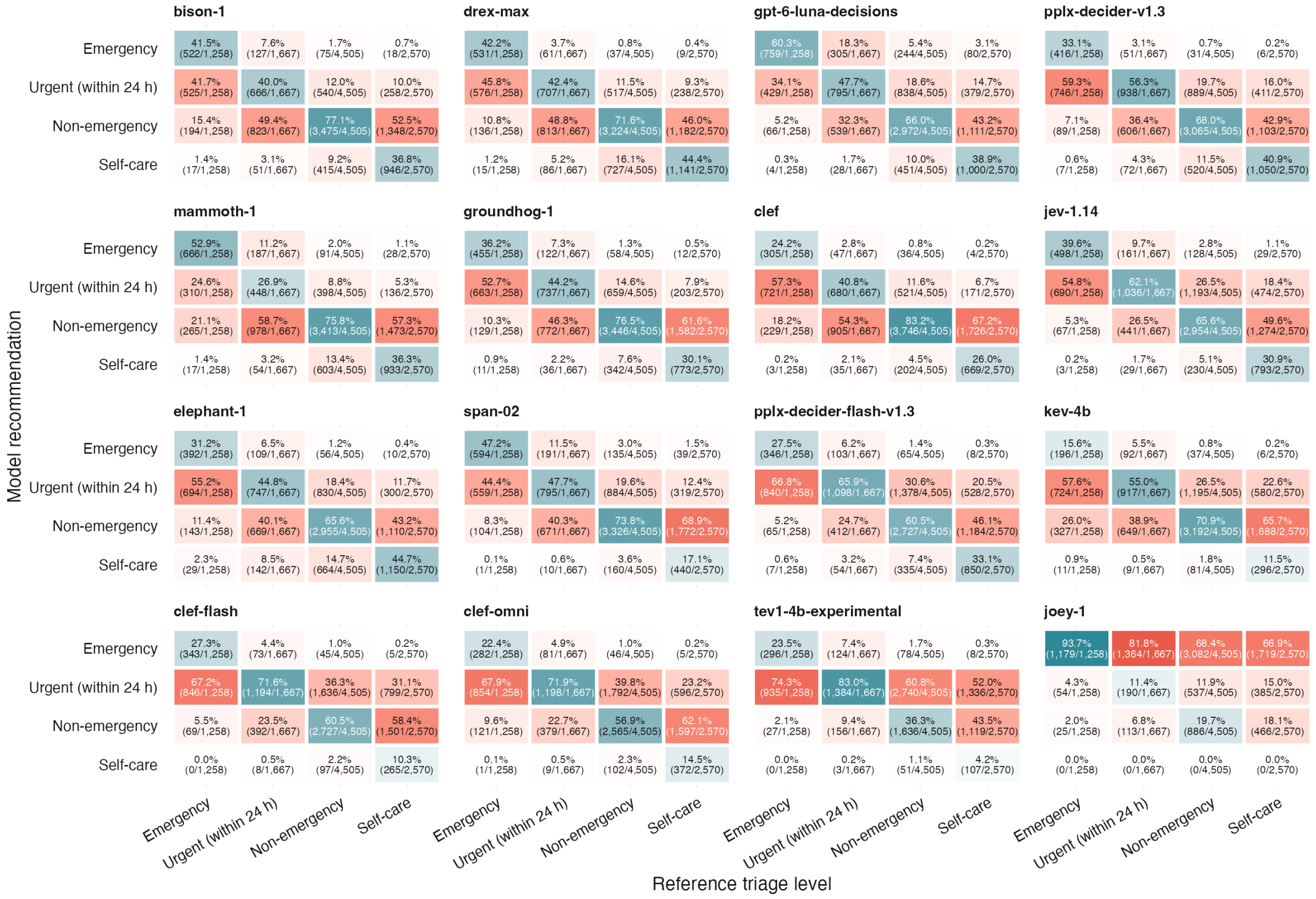

**Figure 4. Confusion matrices per model (real-world cases)**

## Binary Triage Decisions in Real-World Cases

Each model returned a binary triage decision for all 10,000 real-world cases except gpt-6-luna-decisions, which repeatedly declined to provide a decision for one case.

When the triage decision was split into a decision tree with three binary questions, groundhog-1 showed the highest agreement with the SAA, matching its triage level in 56.5% (5,653/10,000; 95% CI 55.6–57.5%), whereas joey-1 showed the lowest agreement (18.1% (1,807/10,000; 95% CI 17.3–18.8%)). Compared with the direct four-option question on the same cases, the decision tree had a higher agreement for 5 of 16 models (from +2.4 (95% CI +1.6 to +3.2) percentage points for groundhog-1 to +5.8 (95% CI +4.7 to +6.8) for kev-4b), a lower agreement for 7 (from -1.8 (95% CI -2.9 to -0.7) percentage points for clef to -9.7 (95% CI -10.6 to -8.8) for clef-omni), and made no difference for 4 models (drex-max, bison-1, pplx-decider-v1.3 and jev-1.14), see Table 3. When only the first two questions were used, resulting in three triage levels (self-care, non-emergency, emergency), groundhog-1 showed the highest agreement with the SAA, matching its triage level in 69.2% (6,915/10,000; 95% CI 68.2–70.1%), whereas joey-1 showed the lowest agreement (23.8% (2,380/10,000; 95% CI 23.0–24.6%)). Compared with the direct three-option question on the same cases, the decision tree led to higher agreement with the SAA for 5 of 16 models, lower agreement for 7 and made no difference for 4, see Appendix Table S4.

When assessing emergencies, sensitivity ranged from 17.1% (215/1,258) for kev-4b to 95.5% (1,202/1,258) for joey-1, and specificity from 19.1% (1,671/8,742) for joey-1 to 98.5% (8,615/8,742) for kev-4b. When deciding between a physician visit and self-care afterwards, sensitivity ranged from 61.8% (3,816/6,172) for elephant-1 to 99.5% (6,143/6,172) for joey-1, and specificity from 1.3% (34/2,570) for joey-1 to 75.7% (1,945/2,570) for mammoth-1. At the third step, deciding between urgent care and regular non-emergency care, sensitivity ranged

from 33.8% (563/1,667) for elephant-1 to 96.2% (1,604/1,667) for joey-1, and specificity from 12.5% (563/4,505) for joey-1 to 91.3% (4,113/4,505) for mammoth-1, see Figure 5.

## Table 3. Agreement of the decision tree with the SAA

| Model | Decision tree<br>% (95% CI)<br>[agreeing/n] | Emergency | Urgent | Non-emergency | Self-care | Direct question<br>(4 options)<br>% (95% CI)<br>[agreeing/n] | Difference<br>(percentage points;<br>95% CI) |
|---|---|---|---|---|---|---|---|
| groundhog-1 | 56.5% (55.6–57.5%)<br>[5,653/10,000] | 58.7%<br>[738/1,258] | 32.0%<br>[533/1,667] | 69.2%<br>[3,117/4,505] | 49.2%<br>[1,265/2,570] | 54.1% (53.1–55.1%)<br>[5,411/10,000] | +2.4<br>(+1.6 to +3.2) |
| drex-max | 56.1% (55.1–57.1%)<br>[5,611/10,000] | 59.3%<br>[746/1,258] | 29.5%<br>[492/1,667] | 57.6%<br>[2,594/4,505] | 69.2%<br>[1,779/2,570] | 56.0% (55.1–57.0%)<br>[5,603/10,000] | +0.1<br>(-0.8 to +1.0) |
| span-02 | 55.9% (54.9–56.8%)<br>[5,585/10,000] | 56.0%<br>[704/1,258] | 27.7%<br>[462/1,667] | 65.3%<br>[2,942/4,505] | 57.5%<br>[1,477/2,570] | 51.5% (50.6–52.5%)<br>[5,155/10,000] | +4.3<br>(+3.2 to +5.4) |
| bison-1 | 55.2% (54.2–56.2%)<br>[5,519/10,000] | 52.4%<br>[659/1,258] | 37.4%<br>[624/1,667] | 60.6%<br>[2,728/4,505] | 58.7%<br>[1,508/2,570] | 56.1% (55.1–57.1%)<br>[5,609/10,000] | -0.9<br>(-1.8 to 0.0) |
| pplx-decider-v1.3 | 55.0% (54.1–56.0%)<br>[5,504/10,000] | 58.5%<br>[736/1,258] | 31.0%<br>[517/1,667] | 55.6%<br>[2,503/4,505] | 68.0%<br>[1,748/2,570] | 54.7% (53.7–55.7%)<br>[5,469/10,000] | +0.4<br>(-0.7 to +1.4) |
| pplx-decider-flash-v1.3 | 54.3% (53.4–55.3%)<br>[5,434/10,000] | 43.5%<br>[547/1,258] | 31.1%<br>[519/1,667] | 63.4%<br>[2,858/4,505] | 58.8%<br>[1,510/2,570] | 50.2% (49.2–51.2%)<br>[5,021/10,000] | +4.1<br>(+3.1 to +5.2) |
| mammoth-1 | 52.7% (51.8–53.7%)<br>[5,274/10,000] | 57.2%<br>[719/1,258] | 22.3%<br>[371/1,667] | 49.7%<br>[2,239/4,505] | 75.7%<br>[1,945/2,570] | 54.6% (53.6–55.6%)<br>[5,460/10,000] | -1.9<br>(-2.9 to -0.8) |
| jev-1.14 | 52.5% (51.5–53.5%)<br>[5,247/10,000] | 59.4%<br>[747/1,258] | 56.6%<br>[944/1,667] | 46.7%<br>[2,104/4,505] | 56.5%<br>[1,452/2,570] | 52.8% (51.8–53.8%)<br>[5,281/10,000] | -0.3<br>(-1.2 to +0.6) |
| clef | 52.2% (51.2–53.2%)<br>[5,219/10,000] | 47.8%<br>[601/1,258] | 64.4%<br>[1,074/1,667] | 47.1%<br>[2,121/4,505] | 55.4%<br>[1,423/2,570] | 54.0% (53.0–55.0%)<br>[5,400/10,000] | -1.8<br>(-2.9 to -0.7) |
| kev-4b | 51.8% (50.8–52.8%)<br>[5,179/10,000] | 17.1%<br>[215/1,258] | 42.1%<br>[702/1,667] | 59.5%<br>[2,679/4,505] | 61.6%<br>[1,583/2,570] | 46.0% (45.0–47.0%)<br>[4,601/10,000] | +5.8<br>(+4.7 to +6.8) |
| elephant-1 | 50.1% (49.2–51.1%)<br>[5,014/10,000] | 36.1%<br>[454/1,258] | 25.3%<br>[422/1,667] | 49.0%<br>[2,209/4,505] | 75.1%<br>[1,929/2,570] | 52.4% (51.5–53.4%)<br>[5,244/10,000] | -2.3<br>(-3.3 to -1.3) |
| gpt-6-luna-decisions | 49.7% (48.8–50.7%)<br>[4,973/9,999] | 75.5%<br>[949/1,257] | 58.8%<br>[980/1,667] | 40.3%<br>[1,816/4,505] | 47.8%<br>[1,228/2,570] | 55.3% (54.3–56.2%)<br>[5,525/9,999] | -5.5<br>(-6.4 to -4.6) |
| clef-flash | 40.9% (40.0–41.9%)<br>[4,093/10,000] | 43.5%<br>[547/1,258] | 78.9%<br>[1,316/1,667] | 36.3%<br>[1,635/4,505] | 23.2%<br>[595/2,570] | 45.3% (44.3–46.3%)<br>[4,529/10,000] | -4.4<br>(-5.3 to -3.5) |

| Model | Decision tree<br>% (95% CI)<br>[agreeing/n] | Emergency | Urgent | Non-emergency | Self-care | Direct question<br>(4 options)<br>% (95% CI)<br>[agreeing/n] | Difference<br>(percentage points;<br>95% CI) |
|---|---|---|---|---|---|---|---|
| tev1-4b-experimental | 38.7% (37.7–39.7%)<br>[3,870/10,000] | 45.5%<br>[573/1,258] | 74.4%<br>[1,241/1,667] | 34.9%<br>[1,571/4,505] | 18.9%<br>[485/2,570] | 34.2% (33.3–35.2%)<br>[3,423/10,000] | +4.5<br>(+3.7 to +5.3) |
| clef-omni | 34.5% (33.5–35.4%)<br>[3,446/10,000] | 42.5%<br>[535/1,258] | 77.4%<br>[1,291/1,667] | 30.1%<br>[1,355/4,505] | 10.3%<br>[265/2,570] | 44.2% (43.2–45.1%)<br>[4,417/10,000] | -9.7<br>(-10.6 to -8.8) |
| joey-1 | 18.1% (17.3–18.8%)<br>[1,807/10,000] | 95.5%<br>[1,202/1,258] | 7.6%<br>[127/1,667] | 9.9%<br>[446/4,505] | 1.2%<br>[32/2,570] | 22.6% (21.7–23.4%)<br>[2,255/10,000] | -4.5<br>(-5.0 to -4.0) |

| | Step 1: Emergency care vs. no emergency care | | Step 2: Physician visit vs. self-care | | Step 3: Urgent care vs. non-emergency | |
|---|---|---|---|---|---|---|
| | Emergency correctly identified (sensitivity) | Emergency correctly ruled out (specificity) | Physician visit correctly identified (sensitivity) | Self-care correctly identified (specificity) | Urgent care within 24 h correctly identified (sensitivity) | Regular non-emergency care correctly identified (specificity) |
| groundhog-1 | 58.7% (738/1,258) | 94.4% (8,252/8,742) | 86.8% (5,360/6,172) | 49.2% (1,265/2,570) | 48.8% (813/1,667) | 84.6% (3,812/4,505) |
| drex-max | 59.3% (746/1,258) | 97.2% (8,493/8,742) | 70.4% (4,346/6,172) | 69.2% (1,779/2,570) | 39.0% (650/1,667) | 90.5% (4,077/4,505) |
| span-02 | 56.0% (704/1,258) | 94.6% (8,267/8,742) | 82.6% (5,101/6,172) | 57.7% (1,484/2,570) | 43.6% (726/1,667) | 85.6% (3,858/4,505) |
| bison-1 | 52.4% (659/1,258) | 95.2% (8,319/8,742) | 81.6% (5,035/6,172) | 58.7% (1,509/2,570) | 53.6% (894/1,667) | 80.8% (3,640/4,505) |
| pplx-decider-v1.3 | 58.5% (736/1,258) | 96.3% (8,415/8,742) | 70.9% (4,373/6,172) | 68.0% (1,748/2,570) | 42.8% (714/1,667) | 88.0% (3,963/4,505) |
| pplx-decider-flash-v1.3 | 43.5% (547/1,258) | 95.8% (8,371/8,742) | 79.0% (4,874/6,172) | 58.8% (1,510/2,570) | 44.1% (735/1,667) | 86.9% (3,915/4,505) |
| mammoth-1 | 57.2% (719/1,258) | 95.3% (8,335/8,742) | 63.5% (3,920/6,172) | 75.7% (1,945/2,570) | 37.7% (628/1,667) | 91.3% (4,113/4,505) |
| jev-1.14 | 59.4% (747/1,258) | 92.3% (8,073/8,742) | 83.1% (5,129/6,172) | 56.5% (1,453/2,570) | 77.5% (1,292/1,667) | 64.1% (2,886/4,505) |
| clef | 47.8% (601/1,258) | 95.9% (8,381/8,742) | 84.5% (5,216/6,172) | 55.4% (1,423/2,570) | 76.2% (1,271/1,667) | 64.0% (2,884/4,505) |
| kev-4b | 17.1% (215/1,258) | 98.5% (8,615/8,742) | 76.8% (4,738/6,172) | 61.6% (1,583/2,570) | 53.6% (894/1,667) | 79.1% (3,563/4,505) |
| elephant-1 | 36.1% (454/1,258) | 97.4% (8,517/8,742) | 61.8% (3,816/6,172) | 75.1% (1,929/2,570) | 33.8% (563/1,667) | 90.2% (4,062/4,505) |
| gpt-6-luna-decisions | 75.5% (949/1,257) | 87.1% (7,614/8,742) | 88.6% (5,466/6,172) | 47.8% (1,229/2,570) | 88.8% (1,480/1,667) | 54.2% (2,441/4,505) |
| clef-flash | 43.5% (547/1,258) | 96.0% (8,391/8,742) | 96.2% (5,936/6,172) | 23.2% (595/2,570) | 91.5% (1,526/1,667) | 40.3% (1,814/4,505) |
| tev1-4b-experimental | 45.5% (573/1,258) | 94.3% (8,248/8,742) | 96.5% (5,959/6,172) | 18.9% (485/2,570) | 91.8% (1,531/1,667) | 38.6% (1,741/4,505) |
| clef-omni | 42.5% (535/1,258) | 96.1% (8,403/8,742) | 96.7% (5,967/6,172) | 10.3% (265/2,570) | 89.6% (1,494/1,667) | 34.1% (1,538/4,505) |
| joey-1 | 95.5% (1,202/1,258) | 19.1% (1,671/8,742) | 99.5% (6,143/6,172) | 1.3% (34/2,570) | 96.2% (1,604/1,667) | 12.5% (563/4,505) |

**Figure 5. Sensitivity and specificity of each binary question in the decision tree**

## Emergency Recognition Using Red Flags and Chief Complaint

When only demographic information, the chief complaint, the answers to the red-flag questions and the risk factors were provided, joey-1 identified 95.7% (1,204/1,258; 95% CI 94.4–96.8%) of emergencies and ruled out an emergency in 17.2% (1,501/8,742; 95% CI 16.4–18.0%) of the other cases (balanced accuracy 56.4% (95% CI 55.8–57.1%)); kev-4b identified 14.0% (176/1,258; 95% CI 12.1–16.0%) of emergencies and ruled out an emergency in 98.1% (8,580/8,742; 95% CI 97.8–98.4%) of the other cases (balanced accuracy 56.1% (95% CI 55.1–57.0%)); drex-max identified 49.7% (625/1,258; 95% CI 46.9–52.5%) of emergencies and ruled out an emergency in 95.7% (8,369/8,742; 95% CI 95.3–96.1%) of the other cases (balanced accuracy 72.7% (95% CI 71.3–74.1%)). The AUROC was 0.698 (95% CI 0.683–0.713) for joey-1, 0.772 (95% CI 0.757–0.786) for kev-4b and 0.874 (95% CI 0.863–0.884) for drex-max, see Figure 6A.

When the predicted probability was used with a rule-out and a rule-in cutoff, the error rates in the held-out cases stayed close to the targets of 1%, 5% and 10% (missed emergencies 0.9–1.0%, 4.8–5.0% and 9.6–10.1%; non-emergencies flagged as emergency 1.0%, 4.9–5.0% and 9.9–10.0%). With these error margins, the emergency decision could be made after the red-flag questions for 6.0%, 24.8% and 38.8% of all cases with joey-1, for 8.9%, 28.2% and 49.3% with kev-4b and for 20.5%, 51.8% and 71.5% with drex-max, and the assessment could end with an emergency referral for 1.7%, 6.7% and 12.3%; 2.1%, 7.8% and 14.1%; and 4.1%, 10.9% and 17.0%, respectively, see Figure 6C (all 95% CIs within ±1.0 percentage points; cutoffs can be found in Figure 6B and Table S5).

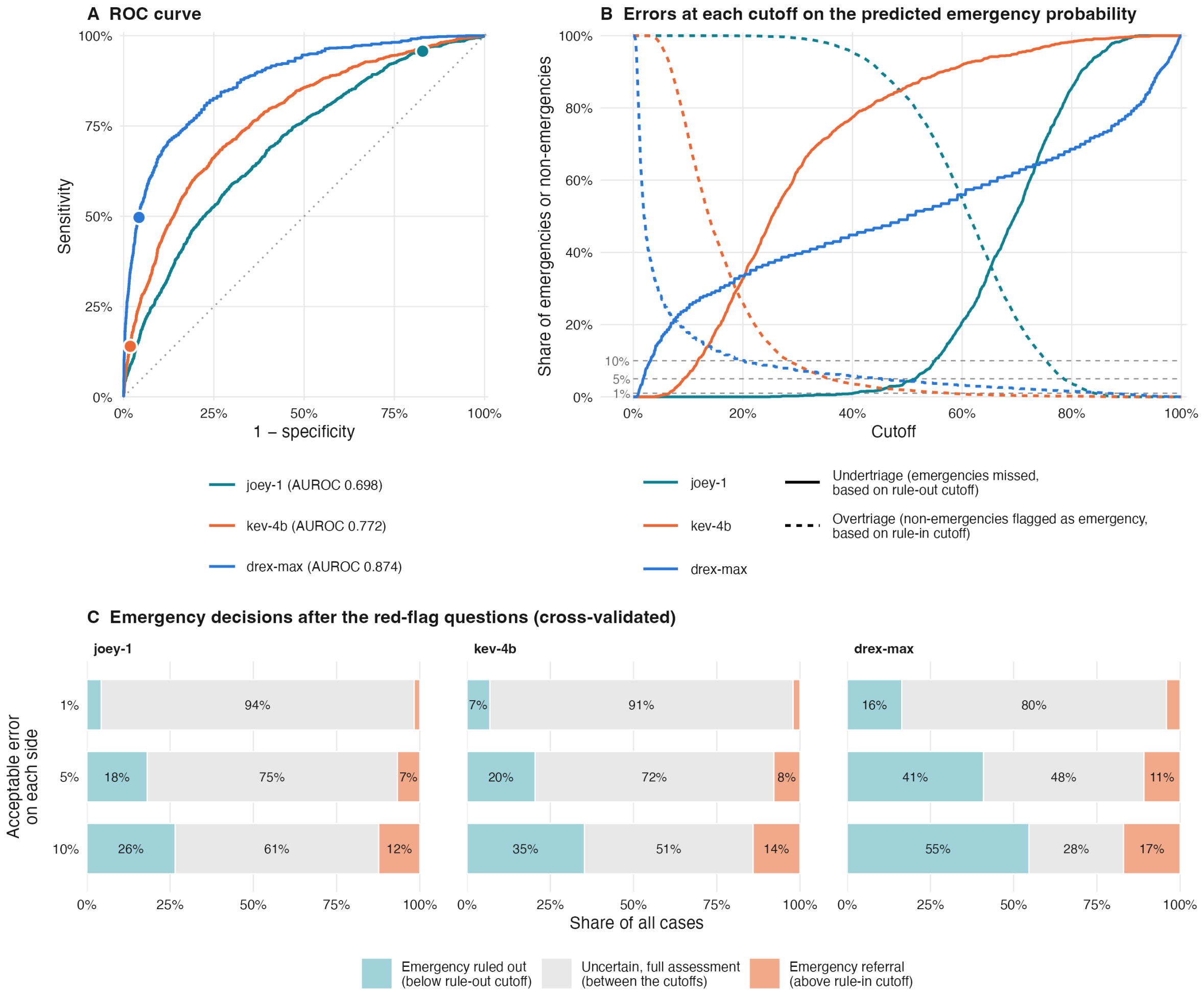

**Figure 6. Emergency decisions based on chief complaint, red flags, and risk factors**

# Discussion

To our knowledge, this was the first study to evaluate the triage accuracy of decision models. We found that decision models are not currently accurate enough to replace existing triage engines in SAAs, but they were promising for one specific sub-decision: recognising emergencies early in the assessment at error margins that can be set in advance. Overall, the best open-weight models matched the best proprietary model with four triage levels, correctly triaging 72% of synthetic cases and agreeing with the SAA in 56% of real-world cases (71% and 55% for the best proprietary model) and exceeded it with three triage levels (80% vs. 73% and 70% vs. 67%, in the same order). Most errors were overtriage errors, whereas emergencies were almost never (0.6%) advised to self-care. Agreement with the SAA also did not consistently improve when the triage decision was split into binary questions: it increased in five models and decreased in seven. Whether assessments could be shortened depended on how the models' output was used. With the reduced input, the models' own decisions either performed better at ruling out emergencies where indicated, but identified less than half of all emergencies overall, or in the case of joey-1, identified almost all emergencies at the cost of very few ruled-out non-emergencies. Since the models returned a probability for each option, however, an acceptable error margin can be set in advance, both for missed emergencies and for false alarms. This approach proved feasible: with such margins, the models could decide after the red-flag questions for a large share of users whether they need emergency care, while unclear cases would continue with the full assessment. In the following, we discuss the most important findings.

Considering our study findings in context, decision models should not be used as standalone triage tools in their current form. Using synthetic cases, the best model triaged 72.1% of cases correctly, which lies within the range reported for LLMs on similar cases (58–76%) [12] and SAAs (12–90%) [5,12]. Conversely, when using real-world cases, agreement with the SAA was even lower, and almost no model recognised more than 60% of emergencies – except for joey-1, which did so at the cost of a very low overall agreement. Decision models thus do not seem to be more accurate than LLMs for triage decisions, which are already used by the public for making care-seeking decisions. They do, however, offer practical advantages for implementation in existing software, because they do not generate output tokens [17]: they generate responses significantly faster, as shown in a recent benchmark study where a decision model needed about 0.3 s per decision compared to a frontier LLM that required 1.2–3.8 s at 50 to 80 times the cost [17]. Further, because some open-weight decision models are much smaller with as few as two or four billion parameters, they can be run locally with less powerful devices and at a lower cost. Their strength may thus lie less in replacing the triage engine of an SAA than in supporting specific sub-decisions within it, such as recognising emergencies early to shorten the assessment, where they could take on tasks otherwise assigned to LLMs at a fraction of the cost.

The most important practical implication is that decision models allow the emergency decision (as such, a sub-decision) to be made early in the assessment, with error margins that can be set in advance for each care setting. When only demographic information, the chief complaint, risk factors and red-flag questions were provided, the model with the highest AUROC could make this decision for about a fifth, half and more than two thirds of all users at error margins of 1%, 5% and 10%, respectively; the remaining users would continue with the full assessment. Because decision models return a probability for each response option, the acceptable share of missed emergencies and the acceptable share of false alarms can be fixed separately: the former through a rule-out cutoff (sensitivity), the latter through a rule-in cutoff (specificity). Most SAAs and LLMs return a single recommendation, so their error margins can be measured but hardly adjusted to a specific setting.

Which error margin is acceptable depends on the care setting. This opens a line of research that is more clinical and regulatory rather than technical: a deliberation on how much accuracy is required and how much risk is acceptable at which touchpoint of care, such as health insurer services, GP practices or emergency departments. To exemplify, in a self-assessment kiosk in an emergency department the primary goal would be that no true emergency is sent away, whereas false alarms carry little cost: users who are not ruled out stay in the emergency department and are triaged by staff as usual. Only the rule-out cutoff would therefore matter, and it could be set for a very small share of missed emergencies (i.e., a very high sensitivity, such as 99%). The specificity at this cutoff then determines the share of non-emergencies who could be offered outpatient care. Even if the specificity at such a strict cutoff is modest, each redirected patient would save resources without additional risk, and because almost half of emergency department patients present without referral and some of them may not even need emergency care, these savings may be high [40,41]. Existing redirection solutions such as digital kiosks in emergency departments have so far shown low undertriage but high overtriage (59–65%) without a confirmed operational benefit [42]. Setting the rule-out cutoff to a defined error margin would make this trade-off explicit and may make an operational benefit more likely. In another setting, regular SAAs in routine use, decision models could instead serve as a second opinion alongside rule-based emergency exits, which end the assessment when specific combinations of red flags and risk factors occur. A model with a predefined error margin might flag emergencies that these rules miss and question exits that are triggered unnecessarily, thereby reducing under- and overtriage. How such a second opinion is best integrated, however, depends on the assessment logic of each system.

Another implication concerns how the triage question is posed: splitting the triage decision into binary questions seems to improve performance only when the individual decisions are clearly separable and only for some models. With both four and three triage levels, the decision tree increased agreement with the reference standard for some models and decreased it for others. This is partly in line with research on LLMs, in which decomposing complex tasks into simpler sub-tasks improved performance for some models [43], and with findings that the recommendations of decision models change when the question format is altered [18]. Although the number of models increasing or decreasing their performance was the same with three and four levels, the models' performance itself partly differed, which might be explained by the third question that distinguished between care within 24 hours and regular care. Sensitivity and specificity for this question varied considerably between models, which suggests that this decision is less clear than the decision for or against emergency care or the decision whether to see a doctor at all. In addition, errors accumulate in a decision tree, as a case can only be triaged correctly if all previous questions were answered correctly. Which decomposition strategies improve accuracy for which models warrants further investigation.

Future studies could build on our results in three directions. First, we only evaluated general-purpose decision models without task-specific adaptation. Since decision models are a new model class that is likely to improve in future releases, and because fine-tuning on triage data might improve the results considerably, the sub-decisions examined here should be re-evaluated with newer and with fine-tuned models. Second, beyond the use case examined here, decision models might support symptom assessments in several other ways: like most SAAs, our evaluation focused on the care need, that is, the level of care a patient requires based on their symptoms, which differs from the patient need, that is, what users actually want from the assessment [44,45]. Decision models could classify free-text input into such patient needs, for example, whether users want care-seeking advice or may just want a sick note or a prescription, for which a full symptom assessment may not be necessary. Established classifications for such encounters, like the ICPC-3 [46], could be used as response options in such cases. Similarly, decision models could extract symptoms from free-text input and assign them to predefined categories to make the symptom assessment shorter. Other shortcuts in the assessment could also be tested, for example, ending an assessment once the probabilities for a triage level no longer change with additional answers, or routing users to specific services such as pharmacies or telemedicine early on. Finally, because decision models return probabilities, it would also be interesting to examine how users deal with probabilities for each option (or the most likely options), instead of receiving a single recommendation only.

## Limitations

First, both datasets have weaknesses regarding their validity. The synthetic cases have a more reliable clinical reference standard that was determined by several physicians, but they do not describe real cases, which limits the external validity of these results. The real-world cases, in contrast, describe actual user interactions and their symptoms, but their reference standard is the triage level assigned by the SAA rather than a clinical decision, which limits internal validity. Results for these cases thus reflect agreement with the SAA, and some disagreements may be errors of the SAA rather than of the models. However, comparing both approaches still yields informative results, as the overall direction was similar in both datasets: overtriage was the most common error, and the same models were among the best and worst performing. Performance was clearly lower for real-world cases, which might be explained by these cases being less clear-cut and more atypical than the textbook-like synthetic cases [24].

Another limitation concerns the source of the real-world cases. All cases stem from a single SAA, which determines the ontology of symptoms and risk factors as well as the questions users are asked. The models therefore only received the information that this SAA considered relevant, which may differ from the information other SAAs or healthcare practitioners collect. However, the structure of chief complaint, red flags, risk factors and follow-up questions is common in SAAs, and this limitation applies equally to all models; thus, the comparison between models and internal validity should not be affected. It does, however, limit how well the absolute accuracy and the emergency shortcut transfer to other systems, which should be examined with data from other SAAs.

Related to this, the real-world cases were provided in German. Responses of LLMs to health questions have been shown to differ across languages, with the best performance in English [47], and the same may apply to decision models. Since the synthetic cases were in English, part of the difference between both datasets might also be explained by language. Performance on real-world cases in English or other languages may thus differ, which should be examined in future studies.

Finally, we used prompts and response options from previous studies [13,28] to make our results comparable with earlier evaluations of LLMs. These may not necessarily be the prompts that lead to the highest accuracy. For LLMs, prompts that reflect human reasoning strategies increased triage accuracy in a previous study [48], and other prompts or response categories may also increase the accuracy of decision models. This point applies to evaluations of language models in general and should be investigated in future studies to identify the best prompts for triage questions.

## Conclusion

The new generation of decision models is useful in identifying emergencies early, but the models should not replace deterministic triage engines in their current form. They may, however, be used to make the emergency decision early in a symptom assessment: because they return a probability for each response option, the acceptable rate of missed emergencies and false alarms can be set in advance, which in this retrospective study allowed the emergency decision to be made early for a large share of users. Which error margin is acceptable needs to be defined for each care setting together with clinicians and regulators. Combined with their output restricted to predefined options, this control over errors may also make decision models easier to certify under the MDR than LLMs. Whether they can shorten other parts of the symptom assessment as well remains an open question.